\pdfoutput=1
\documentclass{llncs}
\usepackage[utf8]{inputenc}
\usepackage{mathtools}
\usepackage{graphicx}
\usepackage[utf8]{inputenc}
\PassOptionsToPackage{hyphens}{url}\usepackage{hyperref}
\usepackage{array}
\usepackage{eurosym}
\usepackage[section]{placeins}
\usepackage{float}
\usepackage{footmisc} 
\usepackage{hyperref} 
\usepackage{booktabs}
\usepackage[dvipsnames]{xcolor}
\usepackage{tikz}
\usepackage{times}
\usepackage{epsfig}
\usepackage{graphicx}
\usepackage{amsmath}
\usepackage{amssymb}
\usepackage{float}
\usepackage{dirtytalk}
\usepackage{booktabs}

\newcolumntype{M}[1]{>{\centering\arraybackslash}m{#1}}
\author{Carlos Rodríguez - Pardo \inst{1,2} \thanks{Worked performed at the University of Edinburgh}, Hakan Bilen \inst{2}}

\institute{
Seddi Labs, Madrid, Spain
\email{carlos.rodriguezpardo.jimenez@gmail.com} \and
School of Informatics, University of Edinburgh, United Kingdom}

\title{Personalised aesthetics with residual adapters}

\begin{document}
\maketitle

\begin{abstract}
 The use of computational methods to evaluate aesthetics in photography has gained interest in recent years due to the popularization of convolutional neural networks and the availability of new annotated datasets. Most studies in this area have focused on designing models that do not take into account individual preferences for the prediction of the aesthetic value of pictures. We propose a model based on residual learning that is capable of learning subjective, user-specific preferences over aesthetics in photography, while surpassing the state-of-the-art methods and keeping a limited number of user-specific parameters in the model. Our model can also be used for picture enhancement, and it is suitable for content-based or hybrid recommender systems in which the amount of computational resources is limited. \\ 
\end{abstract}
\section{Introduction}

The perception of aesthetics in photography, as in many art forms, is typically considered to be subjective. This puts limits on the kind of features that computational models can use in order to predict the aesthetic value of any given picture. There are several factors that influence humans' perception of beauty and aesthetics in art, such as their past experiences, social influences, the situation that surrounds them, their mood or the characteristics of the piece of art itself \cite{Leder2010AJudgments}. The features in which the piece of art can be described with will, therefore, explain only a small part of the preferences of human perceivers over aesthetics in art. Nevertheless, it has been suggested that there is some shared perception of beauty and aesthetics in art \cite{Hayn-Leichsenring2017SubjectivePaintings}, which suggests that there are features that computational models can learn which account for this shared perception of beauty.

The evaluation of aesthetics in photography using machine learning models has gained popularity in the literature in recent years, due to the creation of new annotated datasets and the progress made on computer vision models. The automatic evaluation of aesthetics in photography has many applications, such as automatic image enhancement, the creation of content-based recommender systems or the development of aesthetics-aware image search engines. Most of the progress in this field has been made on building models that can predict a mean aesthetics score of any given picture. Although this approach is certainly powerful, it does not take into account individual preferences over aesthetics in photography, which limits its potential. 

The problem of taking into account subjective preferences on image aesthetics prediction is referred to as \textit{personalized image aesthetics} \cite{Ren2017PersonalizedAesthetics}. Most recent approaches to image aesthetics evaluation have used different deep-learning models, which require a significant amount of annotated data for their training and evaluation. In real-world situations, it is unrealistic to assume that we will have thousands of annotated examples of rated images for any given user. This puts limits on the use of deep learning models for personalized image aesthetics prediction. 

In order to train a machine learning model capable of taking into account individual preferences over aesthetics in photography, an annotated dataset with the identities of the raters of each picture is needed. One example of this kind of dataset is the \textit{FLICKER-AES} dataset, presented by Ren et al. \cite{Ren2017PersonalizedAesthetics}, which contains over 40000 images rated by more than 200 different human raters. Their study provides, along with this dataset (and another, smaller, dataset), a residual-based learning model capable of taking into account user-specific preferences over aesthetics in photography.

We build on their work, and propose an end-to-end convolutional neural network model capable of modelling user-specific preferences with different levels of abstraction, while keeping a reduced number of user-specific parameters within the model. Our method models user-specific preferences by using residual adapters, which were presented in \cite{Rebuffi2018EfficientNetworks,Rebuffi2017LearningAdapters} and have shown success in multi-domain learning. The main difference between our model and Ren et al.'s is that they model user-specific preferences by first training a \textit{generic aesthetics network}, which predicts a mean aesthetic score, and computes a user-specific offset by training a Support Vector Regressor using the predicted content and some manually-defined attributes of the picture as its input; whereas our model embeds the user-specific parameters in different layers of the neural network, therefore allowing the model to find user-specific features with different levels of abstraction, and which do not necessarily depend on the contents and a fixed set of attributes of the pictures. 

Our main contributions are as follows: First, we propose an end-to-end deep neural network architecture capable of surpassing the state-of-the-art results in personalized image aesthetics prediction, while keeping a reduced number of user parameters. Second, we compare different strategies to modelling user-specific preferences over aesthetics in photography using deep learning. Finally, we show how our model can be used for personalized image enhancement, by taking a gradient-ascent approach. For reproducibility reasons, we share our code and trained models in a public repository. \footnote{Please visit \url{https://git.io/fjiKY} for our implementation in PyTorch}.


\section{Related work}
\subsection{Computational aesthetic assessment in photography} The use of computational methods for the evaluation of image aesthetics has gained popularity recently due to at least two factors. First, deep convolutional neural networks (CNN) have created the possibility of learning aesthetic-related features from annotated data. They allow the creation of models that can analyze any picture and predict their aesthetic value, without the need for any annotated data about its contents; and without making use of hand-crafted features. Some examples of the use of CNNs for image aesthetics prediction and related topics can be found in \cite{AppuShaji2016UnderstandingLearning,Lu2014Rapid:Learning,Wang2017DeepAssessment,Chen2017QuantitativeStudy,Hayn-Leichsenring2017SubjectivePaintings,Bianco2016PredictingLearning,Deng2017ImageSurvey,Yu2018Aesthetic-basedRecommendation,DenzlerConvolutionalPerception,Kong2016PhotoAdaptation}. Some of those papers make use of information about the contents of the pictures to improve the predictions of the models. Despite the increasing popularity of deep CNNs for image aesthetics evaluation, there are several studies that use different machine learning algorithms to solve this problem, such as \cite{Datta2006StudyingApproach,Niu2012WhatApproach,Jin2010LearningPhotographs,Vogel2012EvaluatingPortraiture,Jiang2010AutomaticImages,YanKeTheAssessment,Luo2008PhotoSubject,Bhattacharya2010AAesthetics}. Nevertheless, they usually require a significant effort in manually crafting features, which is not a limitation of CNN-based models.

The second factor that allowed the emergence of computational methods for automatic aesthetics assessment in photography is the creation of new and larger datasets. The most popular dataset in the literature of this topic is the \textit{Aesthetic Visual Analysis} (AVA) dataset, which was proposed in \cite{MurrayAVA:Analysis} and which contains around $250000$ pictures, each with an associated histogram of ratings. Nevertheless, there are other datasets that have been used to study this problem, such as \textit{CUHX-PQ} \cite{Tang2013Content-BasedAssessment}, \textit{DPChallenge} \cite{Datta2008AlgorithmicExposition}, or \textit{Photo.Net} \cite{Joshi2011AestheticsImages}.

Even so, none of the studies or datasets mentioned above allows the creation of models that account for user-specific preferences. Examples of studies that actually take into account those kinds of preferences can be found in \cite{Schafer2007,Isinkaye2015,Hong2013,ODonovan2014CollaborativeAesthetics,RotheCVL2016SomePrediction}. However, to our knowledge, the most powerful and innovative study in this field is Ren et al.'s paper \cite{Ren2017PersonalizedAesthetics}, which not only proposes one CNN-based model, but it also publishes two datasets with the information on how each user rated each picture. Their work is important because they show that deep-learning-based models allow the learning of user-specific features, and the two datasets they introduced allow the comparison between models, so that we can empirically analyze which model is more appropriate for this problem. A main limitation of their model, one which we wanted to address, is that the user-specific preferences are learned by a Support Vector Regressor that only takes into account features related to the contents and some other manually-chosen attributes of the images.

\subsection{Transfer learning} One way to overcome the need for huge amounts of data of deep-learning systems is to make use of transfer learning. Usually, transfer learning \cite{Browniee2017ALearning} involves the use of a model that was trained on one task as a starting point to solve another, usually related, task. This method has gained popularity in computer vision problems, as the features learned by CNNs are, to some extent, largely reusable to many different tasks. Closely related to transfer learning, \textit{Multi-domain learning} is concerned with the problem of using a single machine learning model that is capable of solving the same task in different domains. To solve this problem, we can find the solution proposed in \cite{Rebuffi2017LearningAdapters,Rebuffi2018EfficientNetworks}, which is referred to as \textit{residual adapters}. This method consists of the addition of a set of small (usually with a kernel size of 1) domain-specific convolutional layers in parallel to the convolutional layers of a bigger, domain-agnostic, convolutional neural network. The idea is that most of the features in the network are embedded by the domain-agnostic part of the network, and the domain-specific layers compute a small adaptation of the features in the domain-agnostic part of the network. This methodology heavily exploits parameter reuse, and allows the training of deep learning models in situations in which there is only a limited amount of data available for each domain. We believe that those adapters can effectively model user-specific preferences over aesthetics. 

\section{Dataset description}

To train and evaluate our models, we choose the \textit{FLICKR-AES} dataset, which contains around 40500 images rated from 1 to 5 by 210 users. The dataset provides the anonymised identity of the user that gave each rating. Each picture is rated by, on average, around $4.9$ users (standard deviation of $1.87$), with a maximum of 48 and a minimum of 1. The dataset contains pictures with a wide variety of styles and contents, which allows our models to find features that generalize to any style or content. To create our train and test sets, we perform a very similar division to what was done by Ren et al. (we were not able to exactly replicate their train and test division), so that our results are comparable. More precisely, we select the ratings made by 37 users as our test set. Each of those 37 users rated from 105 to 171 images (mean of 137), for a total of 4737. Our train set is composed by the ratings of the remaining 173 workers, and we made sure that the images in the train set did not overlap with the images in the test set, so that our model generalized not only to different users but also to different images. This division was performed so our results could be compared with the baseline model. The ratings were normalized to a 0-mean, unit-variance distribution.

As in \cite{Ren2017PersonalizedAesthetics}, we evaluate our models using the Spearman's ranking correlation ($\rho$) as our main evaluation metric. It is defined as follows: $\rho = 1 - \frac{6 \times \sum_{i=1}^N r_i - \hat{r_i}}{N^2(N-1)}$. This metric is bounded in the $[-1,1]$ range, and measures the correlation between the real ranking $r_i$ of the picture $i$ and its predicted ranking $\hat{r_i}$. Higher values of $\rho$ indicate a better performance of our model.

\section{Experiments}
To design our model, we first create a baseline model that is capable of predicting a mean aesthetic score for each picture, and then we study ways of embedding user-specific preferences to this model. The models are trained using Adam as the learning rule and make use of mini-batches. Images in the training set are preprocessed for each mini-batch as follows: First, as is standard in many computer vision models, they are re-scaled to a 256x256 square picture, then they are horizontally flipped with a probability of 0.5 and a random subsection of 224x224 pixels is cropped (this is the image size that Residual Networks are trained on), then normalized. It can be argued that using data augmentation methods such as random cropping or random flipping can alter the aesthetic value of pictures, but we found that the use of these methods increased the generalization capability of our models. The learning rate was set to $0.001$ for every experiment, and exponentially decayed by a factor of $90\%$ every two epochs. The loss function used was the mean squared error (MSE), measuring the squared difference between predicted and real rating for each picture. 

\subsection{Generic aesthetics prediction model}
We chose a modified version of a ResNet-18 \cite{Alif2017IsolatedNetwork} as our baseline model. Residual networks have shown success in many computer vision tasks, and we chose this shallow version because it was easy to train than deeper versions of this kind of architecture. Instead of randomly initializing the weights of the network, we chose a pre-trained network (trained on the ImageNet dataset). This transfer learning method was also used in \cite{Ren2017PersonalizedAesthetics} and, as was argued before, it can exploit parameter reuse. We added 3 blocks of fully-connected layers to the pre-trained Resnet-18. Each block is composed by 1000 hidden PReLU units ($\alpha=0.25$), a dropout ($p=0.5$) \textit{layer} and a batch normalization layer each. This showed a better performance than using the baseline ResNet-18 model. The final layer outputs 1 value, which is the normalized rating prediction. This network is trained using every picture in the training set and evaluated using the ratings and pictures in the test set.

\subsection{Modelling user-specific preferences}
To test how user-specific preferences can be embedded in the network, we assume two situations: 
\paragraph{10 images per user}
 First, we assume that, for each of the 37 users in the test dataset, we observe 10 ratings. Those 10 pictures are chosen at random, and we fine-tune the bottleneck of our baseline model (the three fully connected blocks) using those 10 images. We perform a 10-fold cross-validation experiment for each user (we choose 10 random images, 10 times) and evaluate on the remaining ratings for each user. We do not perform additional experiments using only 10 images as they are not enough for the training of more complex models. 

\paragraph{100 images per user} In our second set of experiments, we assume that we can observe 100 ratings for each user in the test dataset. Using  3-fold validation (doing a 10-fold configuration was computationally too heavy for those experiments), we perform 3 sets of experiments. First, as in the previous section, we only fine-tune the bottleneck of the network. Second, we fine-tune every layer of the network. This method is computationally heavy and it does not allow to have a reduced number of user-specific parameters. However, it can be seen as a baseline from which to compare other models. Third, we study the use of residual adapters as a way of embedding user-specific preferences in the network. The inclusion of residual adapters in the network allows the learning of user-specific preferences with different levels of abstraction while keeping a smaller number of user-specific parameters than fine-tuning the whole network would need to use.

To study the best way of using the residual adapters for this particular task, we perform 5 different experiments. First, as was suggested in \cite{Rebuffi2018EfficientNetworks}, we add the adapters in parallel to all the 3x3 convolutional layers in the ResNet-18 architecture. Additionally, we also experiment on the addition of these adapters only in the last 3 blocks of convolutional layers in the network (with 128, 256 and 512 kernels per layer). This is motivated by the fact that the first block of layers typically learns low-level features, which should be more generalizable for different users than higher level features, usually learned in deeper layers of the network. Finally, we test a simple method of reducing the dimensionality of the adapters by transforming them to a simple $K \times K_1 \times K$ series of small convolutional layers. The value of $K_1$ is the number of user-specific filters ($K:1 < K$) that we allow the network to learn, and K refers to the number of filters that the adapters receive as an input. Therefore, all the adaptation occurs at the $K_1$ filters. We test three configurations: $K_1 = 1$, $K_1 = \frac{K}{2}$ and $K_1 = \frac{K}{4}$. The smaller the value of $K_1$, the lower the complexity that the adapters will have, which can have a positive influence on the generalization capability of the network. During training time, we only train the bottleneck of the network, as well as the adapters for each particular user. The parameters in the batch normalization \textit{layers} are also trained during training time.

\begin{figure}[H]
\includegraphics[width=\columnwidth]{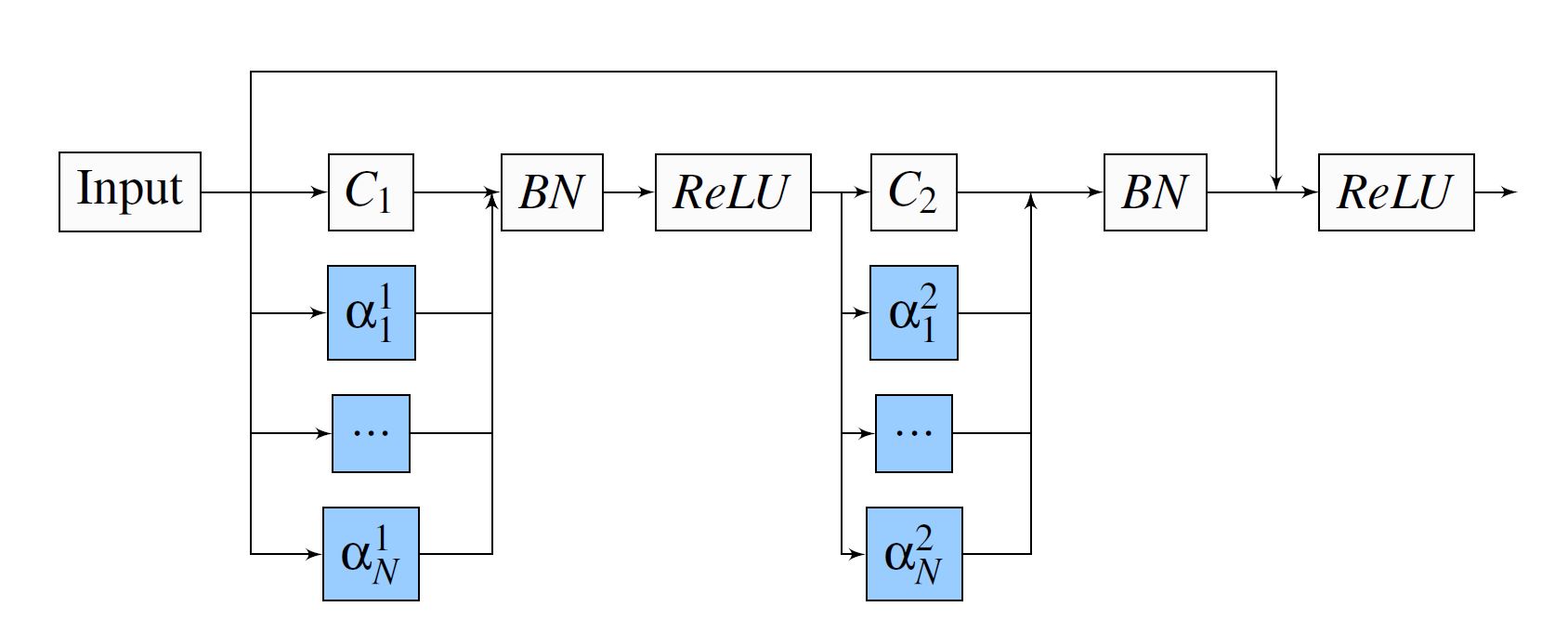}
\centering
\caption{Representation of the residual adapters that were tested in this paper. Each of the $\alpha^i_j$ is a set of K $1 \times 1$ convolutional filters, where $K$ is the number of $3 \times 3$ kernels in the $C_i$ layer, and is uniquely trained for each of the 37 users $j$ in our test set. For our second set of experiments with residual adapters, the $\alpha_i$ layers are actually a series of three layers of $1 \times 1$ convolutional filters: first, a layer of $K$ upcoming feature maps which outputs $K_1$ maps. Second, a layer that receives those $K_1$ feature maps and outputs other $K_1$ maps; and finally a layer that receives $K_1$ maps and outputs $K$ maps. A weight decay of $0.005$ was used on those adapters, to avoid over-fitting and to keep the adapters' initial weights close to 0. It is worth noting that, if the parameters encoded in the adapters are all equal to 0, the network with the adapters is essentially the same as a network without adapters.}
\end{figure}

\section{Results}

Our baseline mean aesthetics prediction network obtained a correlation of $\rho = 0.491$ when predicting the relative ranking of all the pictures of the test set at once. When considering the ratings of each user in the test set separately, and then averaging their correlations, the model obtained a  $\rho = 0.531$. For the personalized aesthetics methods, we randomly selected N ratings for each user and evaluated on the rest of ratings available for that user, averaging the results for the cross-validation experiments, as described in the previous section.

In the experiments where 10 ratings by each user in the test set are available, when fine-tuning the bottleneck of the network, we obtain an average correlation of $0.575$. This shows that, even with a limited amount of information about the preferences of each user, a simple fine-tuning method can learn some adjustments of the weights of the network so that the preferences of each user can be better represented. 

For the experiments in which we assumed that we had the ratings of 100 images for each user in the test set, the results were as follows. We saw that simply fine-tuning the bottleneck of the network was not enough, as it performed significantly worse than other methods $\rho = 0.584$. By fine-tuning all the parameters in the network for each user, we obtained a mean $\rho$ of $0.632$. When using the adapters, we saw that the best configuration was the baseline adapters in parallel to each layer in the network, which obtained a $\rho$ of $0.639$ (and fine-tuning the bottleneck).

\begin{table}[H]
\centering
\begin{tabular}{@{}l||c|c|c|c|c@{}}
Model                   & $\sigma(\rho)$ & Med. $\rho$  & Min. $\rho$    & Max. $\rho$    &  \\ \cmidrule(r){1-6}
Baseline                    & 0.118         & 0.561          & 0.186          & 0.692          &  \\ \cmidrule(r){1-6}
N=10                          & 0.122         & 0.604          & 0.217          & 0.750          &  \\ \cmidrule(r){1-6}
N=100 last layers            & 0.118         & 0.593          & 0.274          & 0.778          &  \\
N=100 all layers             & 0.088         & 0.623          & \textbf{0.458} & 0.828          &  \\ \cmidrule(r){1-6}
N=100, adap. late         & 0.112         & 0.623          & 0.436          & 0.937          &  \\
N=100, adap. all   & 0.115         & \textbf{0.631} & 0.445          & \textbf{0.941} & \\
\cmidrule(r){1-6}
N=100, $K_1=1$        & 0.106         & 0.601          & 0.392          & 0.986          &  \\
N=100,  $K_1=\frac{K}{4}$    & 0.09         & 0.576 & 0.348          & 0.682 & \\
N=100,  $K_1=\frac{K}{2}$    & 0.093         & 0.546 & 0.429          & 0.863 & 
\vspace{5mm}
\end{tabular}
\caption[Comparison of the performance on unseen data of the methods studied]{Comparison of the performance on unseen data for each of the methods studied. In \textbf{bold}, the best result for each category. The value of $N$ is the number of pictures used to learn the preferences of each user in the test set. The experiment named \textit{N=100, last layers}, refers to the experiment in which only the bottleneck of the network was trained, whereas \textit{N=100, all layers} refers to the experiment in which all the network was fine-tuned for each user. All the other $N=100$ experiments use residual adapters, in the ways described in the previous section. When the value of $K_1$ is specified, we refer to the results of the experiments using the \textit{reduced} adapters, whereas the \textit{adap. late} and \textit{adap. all} experiments refer to the baseline adapters positioned in the last 3 blocks of layers and in every block of the network, respectively.}
\end{table}
\vspace*{-\baselineskip}

We confirmed the results of \cite{Rebuffi2018EfficientNetworks}, which showed that removing the adapters from the first block of layers in the residual network did not improve the performance of the model (we obtained $\rho=0.637$) compared to the baseline adapters model. Finally, we could not confirm that the $K \times K_1 \times K$ adapters worked any better than the baseline adapters, so there is no evidence that justifies their use if the goal is to maximize the generalization capability of the model. Nonetheless, we saw that the model in which $K_1 = \frac{K}{4}$ performed considerably worse than when $K_1=1$ or $K_1= \frac{K}{2}$. Interestingly, when $K_1 = \frac{K}{4}$, the results were even worse than the model in which we only fine-tuned the bottleneck of the network. 

\vspace{-\baselineskip} 
\begin{figure}[H]
\includegraphics[width=\columnwidth]{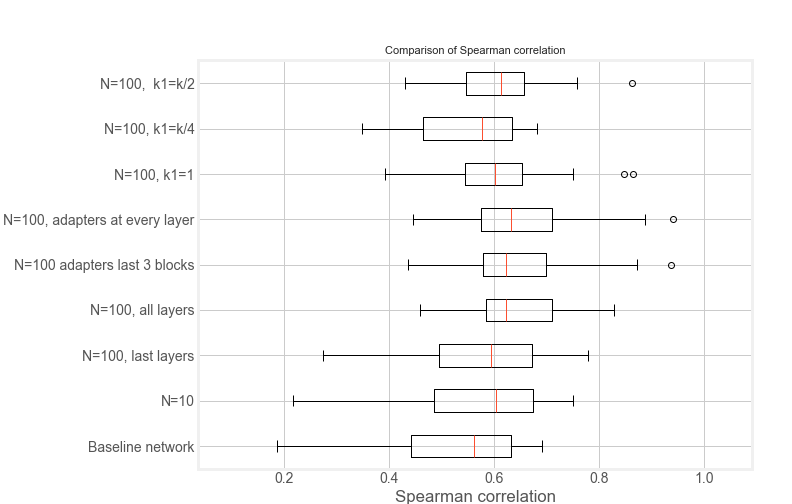}
\centering
\caption{Visualization of the distributions of the Spearman's $\rho$ for each of the 37 workers in the test dataset on unseen data.}
\vspace{-\baselineskip} 

\end{figure}
\vspace{-\baselineskip} 

Despite the fact that the network with the baseline adapters in parallel to each layer has obtained the best mean, median and maximum correlation of all the models that were tested, its results are not significantly different to the results of the network without adapters in which all the parameters in the network were fine-tuned. Nevertheless, we argue that the network with the adapters is preferable as it uses less user-specific parameters, thus reducing the amount of memory needed for each user. In other words, by fine-tuning the whole network, we would need a whole network for each user in the test set, whereas by using the adapters, we would need only one full network for all the users in the test set, and the set of adapters and bottleneck for each of the users in the test set. This difference can be crucial in systems with a large number of users. 

\section{Gradient-ascent picture enhancement}
Given any trained personalized neural network of those specified above, it is possible to use gradient ascent to enhance pictures in a way that the expected rating of the \textit{enhanced} picture is greater than the expected rating of the original picture. To do so, we compute the gradient of the loss function with respect to the input image. By performing this operation: $X_{enhanced} = X_{original} + \epsilon  \times  \nabla_x J(X)$, where $\epsilon$ controls the intensity of the change of the picture and $J(X)$ is the loss function (the prediction of the aesthetic quality of the image), we can obtain an enhanced picture in a considerably fast way. More specifically, the gradient $\nabla_x J(X)$ is computed with respect to the input data and back-propagated to the input image. Then, a small portion ($\epsilon$) of this gradient is added to each of the pixels, which represents the change in each pixel that would maximally increase the expected rating of the picture. One main advantage of this enhancement method is that it does not require a specific architecture for picture enhancement, as it is valid for any deep convolutional neural network that has been trained to predict aesthetic scores in photography. It does not change the style or contents of the picture in any way.

\begin{figure}[H]
\begin{center}
\begin{multicols}{2}
    \includegraphics[width=1.0\linewidth]{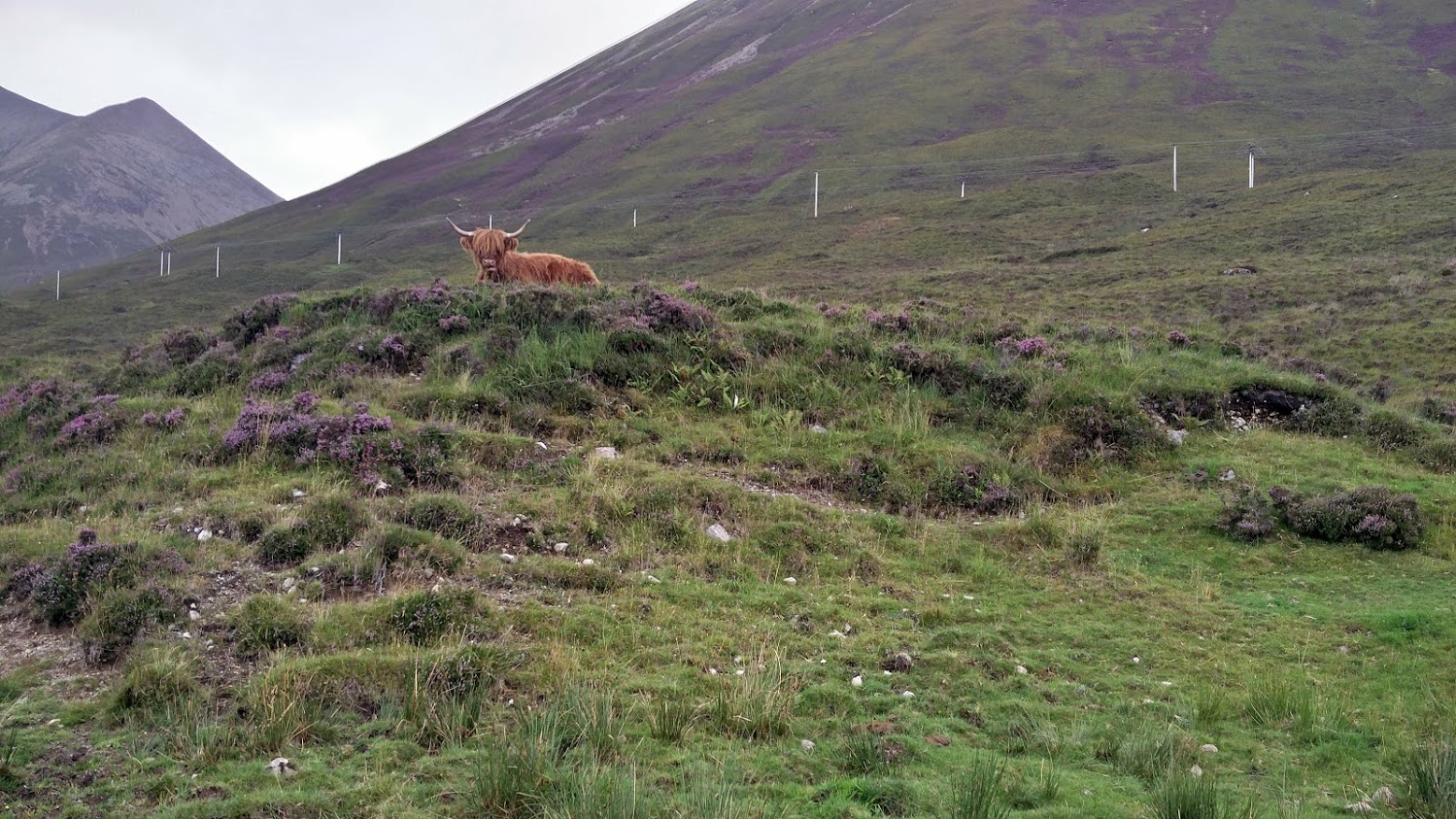}\par 
    \includegraphics[width=1.0\linewidth]{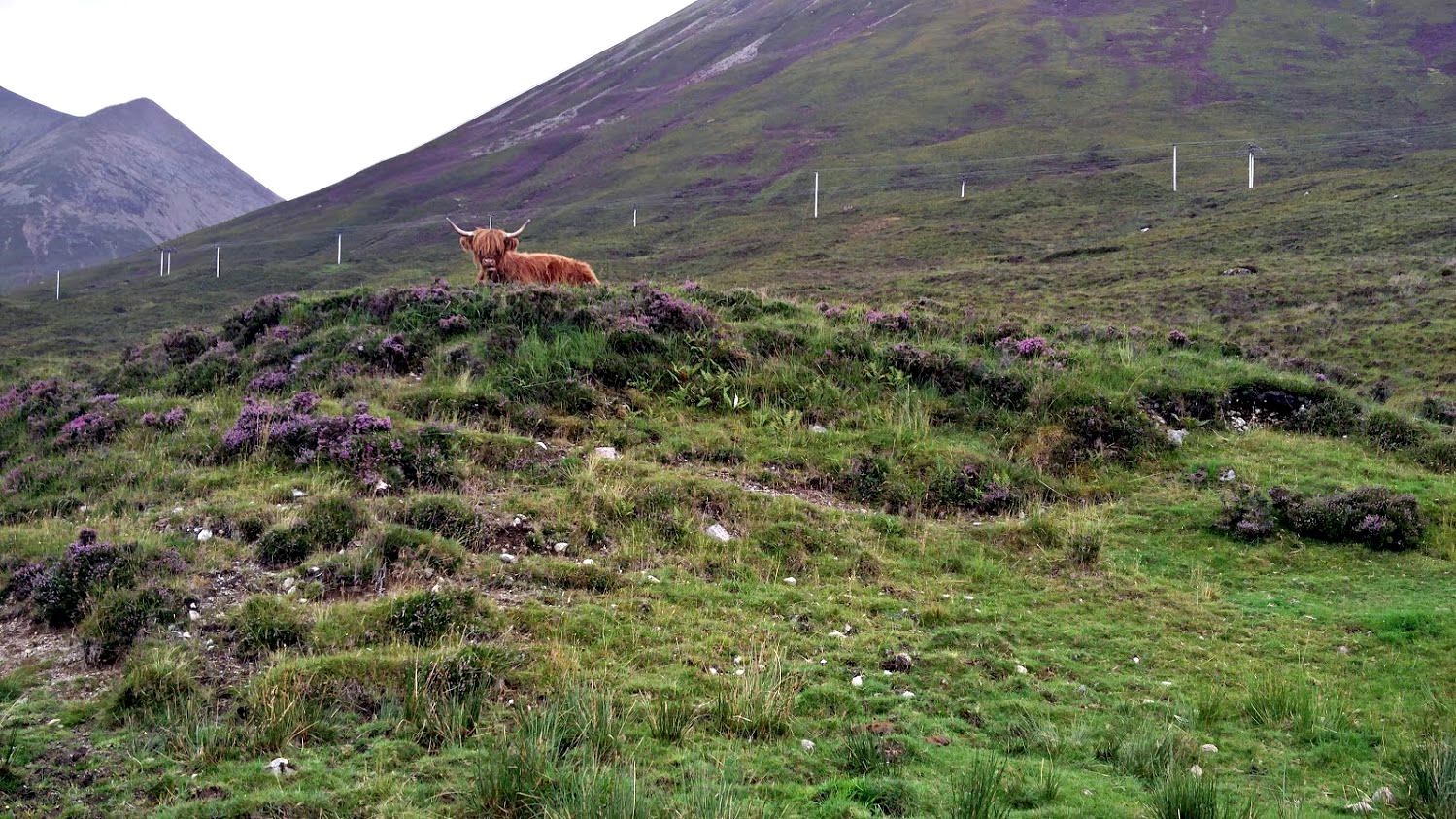}\par 
    \end{multicols}

\begin{multicols}{2}
    \includegraphics[width=1.0\linewidth]{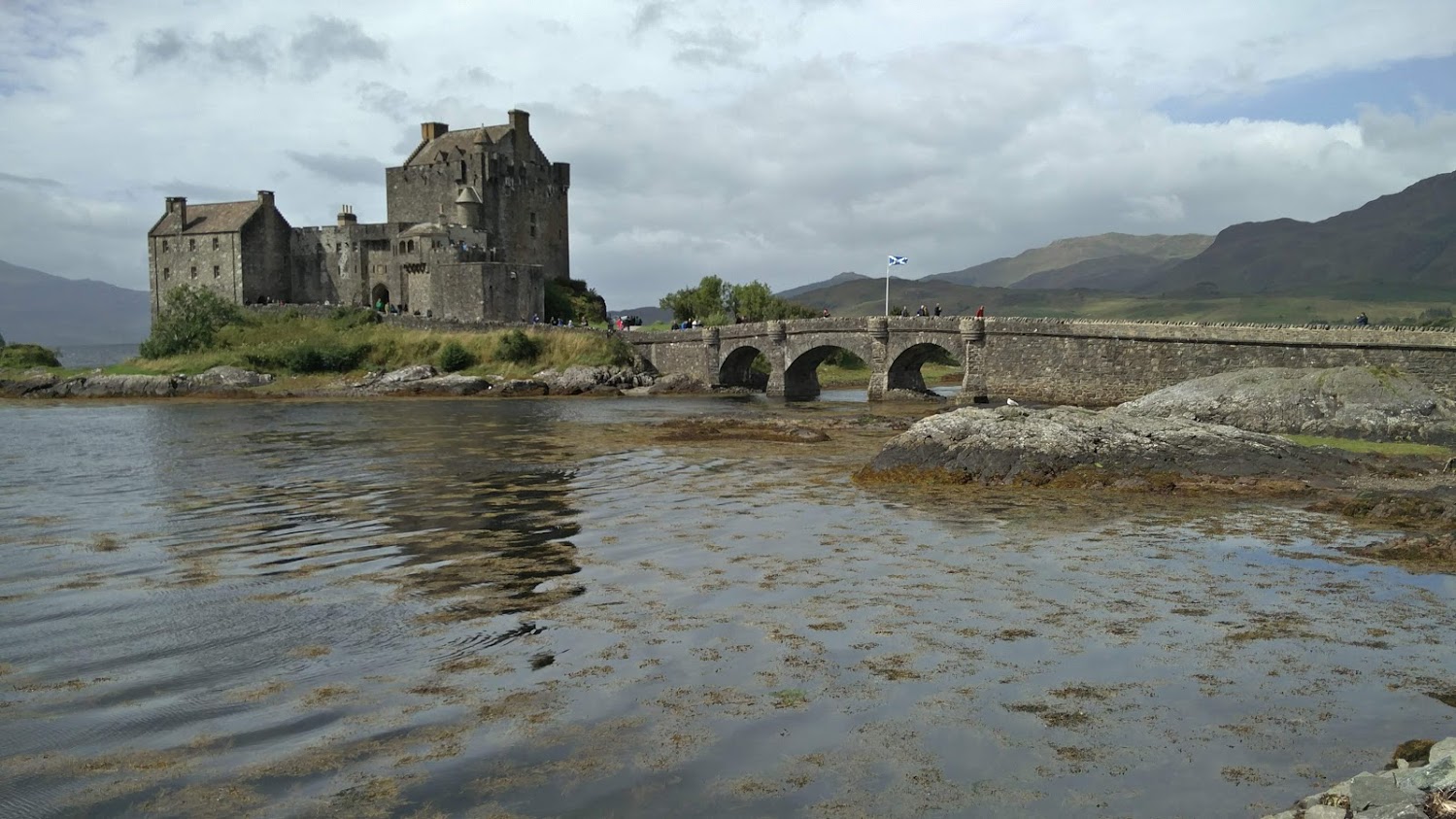}\par
    \includegraphics[width=1.0\linewidth]{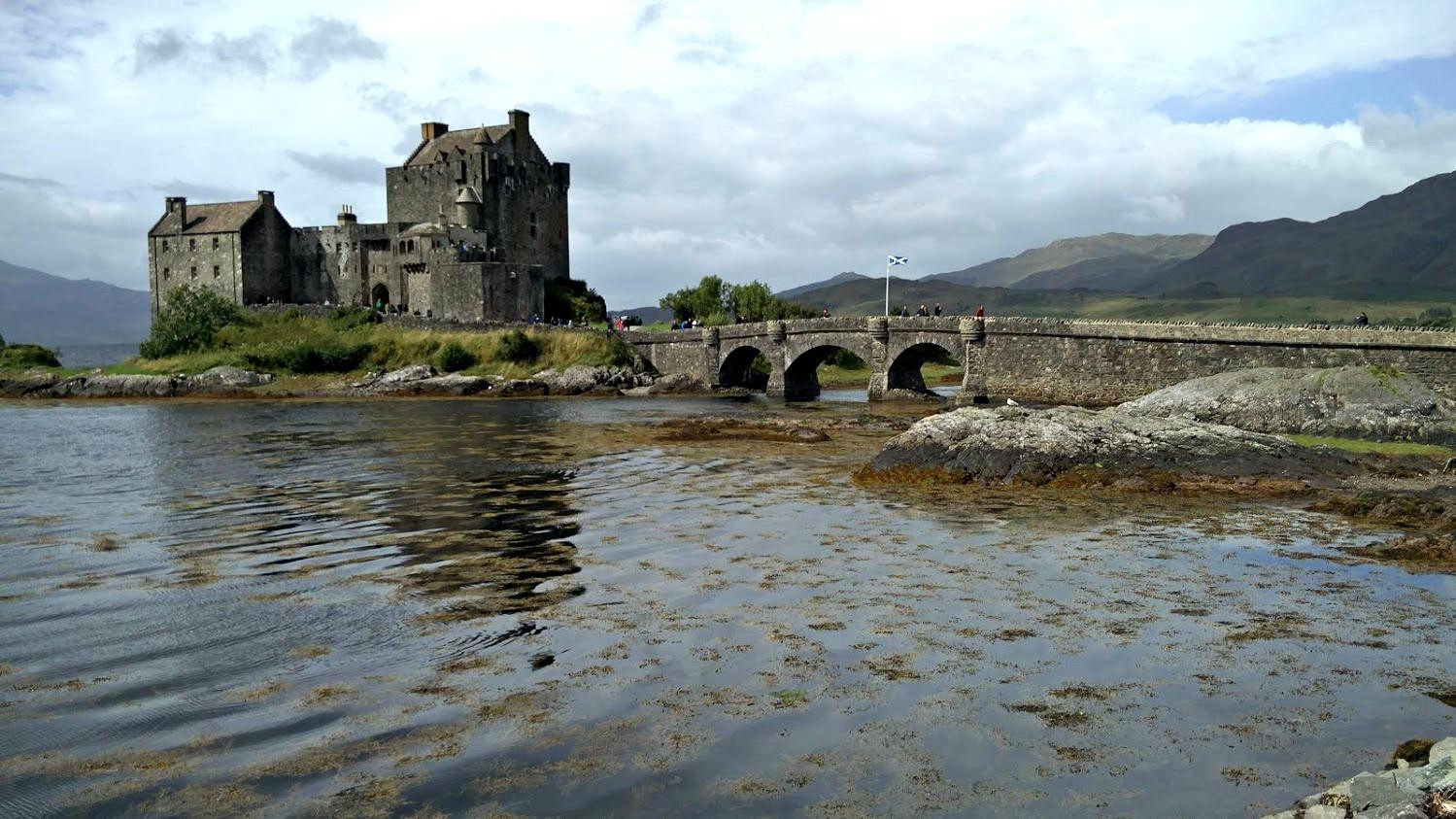}\par
    \end{multicols}
   \label{fig:adversarialenhancement}
\caption[Enhanced images]{Original (left) and enhanced (right) images obtained with our enhancement algorithm and our modified Resnet-18 network. The two original pictures were taken by the authors of this paper using smart-phones, using the HDR configuration. All the enhanced images had a greater expected rating (around $10\%$ bigger) than their corresponding original pictures. The value of $\alpha$ was set to $0.5$ so as to make the differences more visible. It can be seen that the algorithm works for pictures taken under many different lightning conditions. Additional figures are found in our repository. }
    \end{center}

\end{figure}
\vspace{-\baselineskip} 
So that the model could improve pictures of any kind of resolution or proportions (the baseline Res-net only allows 224x224x3 images as input), we added an adaptive average pooling layer \cite{Liu2016AdaptiveClassification} to the start of the network so it adapts any image into the desired input size. Our method can be added to the literature of fast perceptual enhancement using neural networks, a topic studied by other authors \cite{deStoutz2018FastEnhancement}. In sum, our trained network can be used for both evaluating the aesthetic quality of a picture using only a feed-forward pass through the network, and for improving the quality of said picture by back-propagating once the gradient of the loss function, making the enhancement close to real-time (the whole algorithm, takes less than 1 second to process for images of $1920\times1080$ pixels using a GeForce 1050 GPU, and we believe this time can be reduced if the code was optimized). Please see the images above for examples of the enhancements that our method is capable of performing. 

This algorithm can be easily extended so that individual preferences can be taken into account, by using each user's individual personalized network. Consequently, for each user in the dataset, the enhancement that the algorithm will make to each picture will be different. This method creates the potential of personalized picture filters that learn from the past preferences of the users, in a computationally light way.
\section{Conclusions and future work}
In this paper, we have shown that there are deep-learning based methods that are capable of modelling personalized preferences on aesthetic perception in photography, which do not require an explicit modelling of the contents of the pictures or the use of hand-crafted features related to aesthetics or beauty in photography. We have compared different algorithms and ways of modelling said subjective preferences. Our main addition is the proposed residual adapters model, which surpasses the state-of-the-art models in this problem, while keeping a reduced number of user-specific parameters, making our model scalable to real-world applications with a big number of pictures and/or users.  We also confirm the results in the literature \cite{Rebuffi2018EfficientNetworks} regarding the configuration and position of those adapters, and we add some additional information about what way of reducing the dimensionality of those adapters works best. This is valuable for the transfer learning and multi-domain learning literature. The usage of our method in photography recommender systems could be studied in the future.

Our method can be improved in several ways. We used a simple Resnet-18 architecture as the basis for our models. The usage of Inception blocks or other improved architectures can be of interest for this problem, as well as the usage of any optimization algorithms that will be proposed in the future. Data augmentation issues could also be addressed to improve the generalization capabilities of the model, as well as other fine-tuning and other transfer learning methodologies. When new datasets become available, we could also test our hypothesis with more data, as well as improving the network's modelling of aesthetics in photography.

Finally, we also proposed a novel gradient-ascent method capable of performing personalized picture enhancements. Our contribution has been limited to proposing the algorithm and showing preliminary results. However, we believe this method can be studied further and has potential in image quality enhancement problems, which has many applications in research and user applications. 
\bibliographystyle{splncs03.bst}

\bibliography{references}
\end{document}